\documentclass{iopconfser}

\usepackage{microtype}
\usepackage[nolist]{acronym}
\begin{acronym}

\acro{OSL}[OSL]{Optically Stimulated Luminescence}
\acro{TLD}[TLD]{Thermoluminescence Dosimeter}
\acro{APD}[APD]{Active Personal Dosimeter}
\acro{EPD}[EPD]{Electronic Personal Dosimeter}
\acro{MCS}[MCS]{Monte-Carlo Simulation}
\acro{DSR}[DSR]{Dose-Structure Report}
\acro{CT}[CT]{Computed Tomography}
\acro{MRI}[MRI]{Magnetic Resonance Imaging}
\acro{CG}[CG]{Computer Graphics}
\acro{CNN}[CNN]{Convolutional Neural Network}
\acro{PRT}[PRT]{Precomputed Radiance Transfer}
\acro{NeRF}[NeRF]{Neural Radiance Field}
\acro{MLP}[MLP]{Multilayer Perceptron}
\acro{CUDA}[CUDA]{Compute Unified Device Architecture}
\acro{PACS}[PACS]{Picture Archiving and Communication System}
\acro{IR}[IR]{Interventional Radiology}
\acro{VR}[VR]{Virtual Reality}
\acro{AR}[AR]{Augmented Reality}
\acro{API}[API]{Application Programming Interface}

\acro{ICG}[ICG]{Institute of Computer Graphics of the Technical University Braunschweig}
\acro{PTB}[PTB]{Physikalisch-Technische Bundesanstalt}
\acro{ECAP}[ECAP]{Erlangen Centre for Astroparticle Physics}
\acro{SCK}[SCK CEN]{Belgian Nuclear Research Centre}

\acro{WP}[WP]{Work Package}

\end{acronym}

\usepackage{hyperref}
\usepackage[english]{babel}
\usepackage{csquotes}
\usepackage{graphicx}
\usepackage{wrapfig}
\usepackage{esvect}
\usepackage{tikz}
\usepackage{amsmath}
\usepackage{svg}
\usepackage{gensymb}
\usepackage[a4paper]{geometry}
\usepackage{booktabs}

\usepackage{siunitx}
\usepackage[capitalize]{cleveref}
\crefname{paragraph}{Sec.}{Secs.}
\crefname{section}{Sec.}{Secs.}
\Crefname{section}{Section}{Sections}
\Crefname{table}{Table}{Tables}
\crefname{table}{Tab.}{Tabs.}
\Crefname{figure}{Figure}{Figures}
\crefname{figure}{Fig.}{Figs.}
\Crefname{chapter}{Chapter}{Chapters}
\crefname{chapter}{Ch.}{Chs.}

\providecommand{\etal}[0]{\emph{et~al.}}

\usepackage[colorinlistoftodos]
{todonotes}
\definecolor{darkorange}{rgb}{1.0, 0.55, 0.0} 
\definecolor{LightCyan}{rgb}{0.88,1,1}
\definecolor{darkred}{rgb}{0.7, 0.0, 0.0}
\definecolor{forrestgreen}{rgb}{0, 0.36, 0.0}

\def\FINAL{}
\usepackage[normalem]{ulem}
\definecolor{yellow-green}{rgb}{0.4, 0.53, 0.14}
\newcommand{\rev}[1]{\ifx&#1&\else\colorbox{teal!20}{#1}\fi}
\newcommand{\added}[2][]{\ifdefined\FINAL#2\else\rev{#1} \textcolor{yellow-green}{\textbf{#2}}\fi}
\newcommand{\deleted}[2][]{\ifdefined\FINAL\else\rev{#1} \textcolor{red}{\sout{#2}}\fi}
\newcommand{\replaced}[3][]{\ifdefined\FINAL#3\else\rev{#1} \deleted{#2} \added{#3}\fi}

\begin{document}

\title{RadField3D: A Data Generator and Data Format for Deep Learning in Radiation-Protection Dosimetry for Medical Applications}

\author{Felix Lehner$^{1,3}$, Pasquale Lombardo$^{2}$, Susana Castillo$^{3,5}$, Oliver Hupe$^{1}$ and Marcus Magnor$^{3,4,5}$}

\affil{$^1$Physikalisch-Technische Bundesanstalt (PTB), Braunschweig, Germany}
\affil{$^2$Belgian Nuclear Research Centre (SCK CEN), Boeretang, Mol, Belgium}
\affil{$^3$Institute for Computer Graphics, Technische Universität Braunschweig, Braunschweig, Germany}
\affil{$^4$Physics and Astronomy, University of New Mexico, New Mexico, USA}
\affil{$^5$Cluster of Excellence PhoenixD, Leibniz University Hannover, Hannover, Germany}

\email{felix.lehner@ptb.de, lehner@cg.cs.tu-bs.de}

\begin{abstract}
In this research work, we present our open-source Geant4-based Monte-Carlo simulation application, called RadField3D, for generating three-dimensional radiation field datasets for dosimetry. Accompanying, we introduce a fast, machine-interpretable data format with a Python API for easy integration into neural network research, that we call RadFiled3D. Both developments are intended to be used to research alternative radiation simulation methods using deep learning. All data used for our validation (measured and simulated), along with our source codes, are published in separate repositories.\\
\added[]{\url{https://github.com/Centrasis/RadField3DSimulation}\\
\url{https://github.com/Centrasis/RadFiled3D}}
\end{abstract}

\section{Introduction}

Monte-Carlo methods are extensively utilized in the analysis of irradiation scenarios across various domains, including radiation protection~\cite{chatzisavvas_monte_2021}. Radiation protection primarily focuses on preventing unnecessary radiation exposure to humans and the environment. While legislative measures, risk management, and personal dosimeters can mitigate exposure in many areas, certain professions inherently involve inevitable radiation exposure. These include roles in nuclear power plants, waste disposal, and medical treatments such as~\ac{IR}.

In \ac{IR}, medical staff are exposed to non-uniform radiation fields due to their proximity to the patient, making it challenging to accurately assess individual doses. Current personal dosimetry methods are reliable only for uniformly distributed radiation fields, which is not the case in \ac{IR} settings. To address this, previous research has proposed the use of Computational-Dosimetry systems~\cite{oconnor_feasibility_2022}. These systems track the locations and postures of all individuals involved in an \ac{IR} procedure and calculate spatially resolved doses for each person.

However, existing radiation transport simulations lack the speed required for real-time dose calculations, which is also a limitation for Virtual Reality (VR) training software. VR training systems are crucial for reducing radiation exposure during interventions by increasing awareness among medical staff~\cite{rainford_student_2023, fujiwara_virtual_2024}. Despite their importance, these systems lack dosimetrically reliable data for real-time visualization.

To advance radiation protection, there is a need for comprehensive and reliable data sets to allow the development of reliable acceleration techniques for dose calculations. To answer this challenge, we have developed an open-source application capable of calculating spatially resolved, voxelized radiation field distributions using state-of-the-art \ac{MCS} code, specifically Geant4~\cite{agostinelli_geant4simulation_2003}. We will further refer to this software as \textit{RadField3D}. RadField3D uses a specific data format for simulated radiation fields. We implemented this format as an open-source binary file format for storing and loading voxelized radiation fields, accessible from both C++ and Python. \replaced[]{In the further course we call this format}{The format will henceforth be referred to as} \textit{RadFiled3D}.

Our primary innovation is the provision of an open-source application for calculating spatially resolved, three-dimensional radiation fields, validated through laboratory measurements. The validation was performed against measured air kerma rate, as this metric is important for the staff dose estimation later on. The corresponding measurement data will also be published with this paper. Rad\textit{Filed}3D and its related \ac{API} are designed to improve reproducibility and facilitate usability of dosimetric data in multiple use cases. Therefore, the availability of the \ac{API} for Python is significant, given Python's widespread use in data evaluation and machine learning.
\subsection{Monte-Carlo Simulation in Radiation Protection}
In principle, the technique of using \ac{MCS} to calculate the radiation transport of complicated measurements is already widely spread in various disciplines of dosimetry from radiation protection to medical physics. Therefore, there is already a set of well understood and tested general purpose \ac{MCS} toolkits for radiation transport such as Geant4~\cite{agostinelli_geant4simulation_2003}, MCNP~\cite{werner_mcnp_2018}, or EGSnrc~\cite{canada_egsnrc_2021}. As these \added{toolkits} work for a huge set of different situations, a fine tuning of the \acs{MCS} parameters for each specific use-case is required. There are frameworks built upon those general-purpose toolkits, like GATE~\cite{jan_gate_2004} -- based on Geant4 -- already in use for the dosimetry of patients during tomography or nuclear medicine.

Unfortunately, there is a lack of specifically validated \ac{MCS} applications for the spatially resolved simulation of radiation fluence distributions in the space around the patient during interventions. Such an application would be needed for the computational radiation protection dosimetry of staff as it was proposed by prior research projects like the PODIUM project~\cite{oconnor_feasibility_2022}.

\subsection{Computational Dosimetry}
Real world measurements in \ac{IR} situations using \acp{APD} have already demonstrated the complexity of present radiation fields and locally high dose rates to professionals~\cite{hupe_determination_2011}. Those measurements also implied that the usage of \acp{APD} at one single position on the body is likely to under- or overestimate the received doses to the person. During the PODIUM project, the researchers developed a camera-based tracking system coupled with a \ac{MCS} application based on the MCNP~\cite{werner_mcnp_2018} toolkit. Their simulation segmented the room into frustums projected from the surface of a sphere placed around the patient in the isocenter of the scene~\cite{abdelrahman_first_2020, vanhavere_d9121_2020}. Each event was scored as it crossed one of the surface segments of the sphere and effectively projected it along the normal of that surface segment through the scene. Based on these frustums, they validated their simulation against measurements of \acp{APD} worn by medical personnel during real interventions where they tracked the locations and postures of the staff.

\subsection{Acceleration Techniques for Monte-Carlo Simulation}
As \ac{MCS} is several magnitudes too slow for use in a real-time scenario~\cite{vanhavere_advances_2022}, there are on-going efforts to improve the speed of the calculation by substituting single parts or even the whole algorithm of the \ac{MCS} with a neural network~\cite{arias_radiation_2023}. 
Other works denoise prematurely aborted \ac{MCS} calculations to estimate result of the fully performed \ac{MCS}~\cite{bai_deep_2021, peng_mcdnet_2019}. To investigate such approaches properly, there is a need for appropriate datasets. As the datasets of those research works are often not published or not properly validated, we concluded that there is the need for open source datasets and dataset generators.

\subsection{Radiation Field Data Formats and Representations}
Moreover, we concluded that there is a need for a specific data format for such use-cases, even though there are already various dataformats present. For example, there are general purpose formats like the binary file format that belongs to the Root data analysis framework~\cite{root_team_root_nodate} which is maintained by the Geant4 developers. That file format allows the creation of custom tree shaped data structures based on the personal needs. This results in various structures for each developed \ac{MCS} application, even in the same field, making it difficult to reuse data calculated by different researchers. Further, other toolkits like MCNP export into plain text-based files, which are neither efficiently readable nor are they easily understandable. Another possibility for storing voxelized radiation data is the use of the commercially widespread DICOM~\cite{Parisot1995,national_electrical_manufacturers_association_digital_2024} format for the storage of \ac{CT} -- or \ac{MRI} -- scans. Here, we have the issue that this format is too specific, as its purpose is the storage of medical procedures with a defined set of measurands. Also, the reproducibility is not supported for simulated data, as the metadata is defined for existing machines with physical properties that do not apply to simulations. Therefore, we propose a simple, binary file format with a fixed structure that incorporates mandatory metadata following the guidelines of the research work of Sechopoulos~\etal\ on documenting radiation transport simulations~\cite{sechopoulos_records_2018}.
\section{Materials and Methods}
In this research work, we developed a Monte-Carlo radiation transport simulation utilizing the Geant4~\cite{agostinelli_geant4simulation_2003} toolkit, designed to accurately model the interaction of radiation within various materials and environments. To validate the accuracy of our fluence calculations, we employed a statistical error measurement technique based on the incremental evaluation of variances, which was applied to all energy bins of all measured photon fluences.

To facilitate the integration of our simulation data with machine learning algorithms, we designed a robust yet customizable binary file format, leveraging on PyBind11~\cite{wenzel_pybind11_2016} to create Python bindings. This format is organized in structured channels and layers, preserving units and statistical errors for each layer, thereby ensuring the integrity and usability of the data for machine learning applications.

To validate RadField3D as a reliable source of ground truth data, we conducted a series of measurements in radiation fields that are in accordance with the ISO 4037-1~\cite{iso_iso_2019}. Those measurements were performed at three distinct distances around two different phantoms using the H-100 radiation quality to maximize the detector response in the scattered radiation field. The experiments were carried out using PTB's calibration X-ray facility, that offers a \(4\,\%\) uncertainty towards absolute personal equivalent doses for the energy range of \SI{50}{\keV} to \SI{300}{\keV}~\cite{bipm_x-ray_2024}. \deleted[R2C1]{The comparison between our simulated data and measurements was done to test the accuracy and reliability of our simulation, in order to establish it as a solid foundation for further research in radiation protection and machine learning.}

\subsection{Simulation and Voxelization Method}
At the core, the simulation framework of RadField3D is built on top of the Geant4~\cite{agostinelli_geant4simulation_2003} framework. RadField3D integrates some optimization aimed at improving transport data collection. We only interfere minimally with the standard particle transport process of Geant4. A physics list was selected to best fit the use-case of the medical applications. In particular, we combined the \textit{QGSP\_BIC\_HP} physics list with the \textit{G4EmStandardPhysics\ Option4} option. Then, RadField3D registers a \textit{G4UserSteppingAction}, which tallies all the segments of the simulated particle trajectories. On each step, RadField3D automatically categorized particle segments in three separate components, those being \textit{Beam}, \textit{Patient}, and \textit{Scatter}.

\begin{wrapfigure}{r}{5cm}
    \includegraphics[scale=0.43]{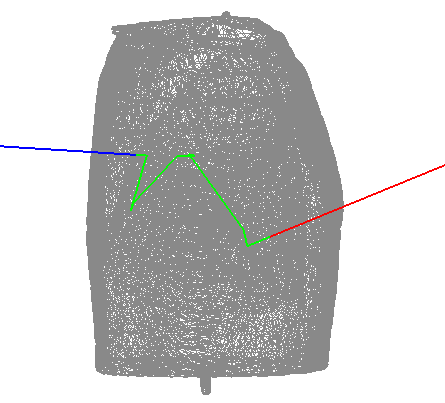}
    \caption{Wireframe render of one single photon track, crossing a scatter phantom geometry, with colored track-components. Red: Direct beam component (\textit{Beam}). Green: Scatter component inside the patient (\textit{Patient}). Blue: Scatter component outside the patient (\textit{Scatter}).}\label{fig:TrackComponents}
\end{wrapfigure}

The components are illustrated in \Cref{fig:TrackComponents}. For our experimental validation, we focused solely on the \textit{Beam} and \textit{Scatter} components. To score the voxel state for each of these components, we retrieved the start and end locations of each step. RadField3D identified which voxels lay on the path of the photon, and at each transition from one voxel to another, it scored the entrance energy. We propose to score basically three different values per voxel: the percent of photons that hit the voxel, the probability distribution of photon energies, and the dominant direction component of the entrance trajectory. Using this information, we can easily compute the actual dose or doserate later on, when the actual measurand has at least one known point. As the secondary electrons in air do not contribute to the detector response directly, we only score photons in the voxels. Nevertheless, we include these electrons in our simulation process until they either lead to the emission of another photon, or their kinetic energy depletes.

In RadField3D, a voxelization subdivision algorithm is used to discretize the scoring of the path of scatter photons. The simulation itself uses continuous coordinates for the geometry and the paths. Therefore, we employed the open asset importer library \textit{assimp}~\cite{noauthor_assimp_2023} and the Geant4 tessellation process to allow RadField3D to load of the mesh geometry from multiple geometry definition formats. For our research work, we used the Wavefront OBJ-file format as well as the Autodesk Filmbox FBX-file format to load the mesh in RadField3D. As those classical file formats do not directly define volumes and materials, we paired the mesh files with description files containing material information, location offset, rotation, and scaling vectors. These description files are encoded as JSON-Documents. The description files also contain information about the nesting of meshes and volumes inside each other, which is used to allow the stacking of different tissues in the phantom which can be declared in those as well.

The training of machine learning models requires very large datasets. To drastically improve the simulation speed and allowing the building of large datasets, we do not employ analytical line-cube or line-triangle intersection tests as it is computationally intensive. Instead, we sample points on the line segment within a maximum distance of \(0.5\) times the minimal voxel extent along each dimension. With this test, RadField3D can efficiently identify intersected voxels while maintaining the same accuracy of a full analytical test and has virtually no risks of subsampling the voxel grid.

 To model the X-ray tube emission, RadField3D uses pre-calculated X-ray Tungsten spectra including realistic combinations of filters and anode polarizations. The library of spectra was generated in the form of a CSV-file using SpekPy~\cite{poludniowski_technical_2021} and is used by RadField3D to sample photon energy probabilities. Further, RadField3D supports radiation field shapes: a simple cone that can be parameterized by an opening angle, and a pyramid that can be parameterized by edge lengths of a rectangle at a defined distance.

For our validation and experiments we used a cubic field shape of \(1\,\mathrm{m}^3\) divided into cubic voxels with dimensions 
\(2\,\mathrm{cm} \times 2\,\mathrm{cm} \times 2\,\mathrm{cm}\), resulting in a grid of \(50 \times 50 \times 50\) voxels per layer.

\subsubsection{Statistical Error}
In order to get a consistent metric for automatically estimating the optimal number of particles in each simulation, we tallied the normalized variance of each energy bin in the photon energy distribution within each voxel. This variance is continuously recalculated over the course of the simulation.

As the photon energy distribution \(p(E_\mathrm{\gamma})\) of each voxel converges against the true energy distribution as more particles are traced, the variance of all bins strives towards zero. Additionally, to get the relative statistical error in the range \(\epsilon_{rel} \in [0..1]\), we normalize the variances to the maximum variance which would occur if the histograms bin values are all uniformly sampled between zeros and ones. This maximum variance would be \(0.25\). Afterwards, we aggregate the bin-wise variances to a single \(\epsilon_{rel}\) per voxel by calculating the mean relative statistical error for all of its bins. Finally, as an estimate of the overall error for each radiation field, we select the \(\epsilon_{rel}\) of that voxel where \(95\,\%\) of all other voxels have a lower \(\epsilon_{rel}\) than that. This criterion provides a consistent termination condition for the simulation of one single radiation field.
The formula for the \(\epsilon_{rel}\) per voxel is the following:\\

\begin{equation}
    \begin{aligned}
        \epsilon_{rel} &= \frac{\sigma^2}{0.25} \\
        \sigma^2 &= \frac{\sum_{i=1}^n{(x_i - \mu)^2}}{n}
    \end{aligned}
\end{equation}

In order to further reduce the calculation effort as well as the memory consumption, we do not use the classical formulation of statistical variance, as we do not want to store each step value for each bin of each voxel. Instead, we use a more memory efficient approach where we calculate mean and variance incrementally. To perform this calculation, we employ Welford's variance online algorithm~\cite{welford_note_1962} which allows us to update the variance on a stream of values. In order to enhance the meaningfulness of the bin-wise variance, we update the calculation of the variance at every $50^{th}$ photon per voxel. That way, we guarantee observable changes in the histograms and a bigger variance, if the voxels histogram is still unstable during the first part of the simulation. To further reduce the computing overhead, a second condition is used where we restrict the evaluation of all voxels statistical error and the evaluation of the simulation termination condition to every $50000^{th}$ photon.

\subsubsection{Conversion of simulated data to absolute measurands}
In order to validate the simulation data of RadField3D against measurements, we compared calculated air kerma rates to measured air kerma rates \(\dot{K}_{\mathrm{air}}\) at the positions of the 
measurements. Therefore, we multiply the 3D tensor of the photon energy probability distribution \(p(\mathbf{E}_\mathrm{\gamma})\) 
with the 3D tensor of the fraction of photon hits per voxel and primary photon \(\overline{\mathbf{F}}_{\mathrm{\gamma}}\) to get the relative energy distribution entering each voxel. Then, we calculate the energy transmission rate \(\overline{\mathbf{E}}_{\mathrm{\gamma}}^{\mathrm{tr}}\) by multiplying the entrance tensor with the energy transmission coefficients \(\vec{\mu}_{\mathrm{tr}}\). The transmission coefficient is approximated by calculating the mean attenuation over each energy bin. Subsequently, we can retrieve the tensor \(\dot{\mathbf{K}}_{\mathrm{air}}\) by calculating the weighted integral of the distribution for each voxel \(\overline{\mathbf{E}}_{\mathrm{\gamma}}^{\mathrm{tr}}\) and its mean energy bin edges \(\mathbf{\bar{E}}_{\mathrm{\gamma}}^{\mathrm{bin}}\).

\begin{equation}
    \begin{aligned}
        \overline{\mathbf{E}}_{\mathrm{\gamma}}^{\mathrm{tr}} &= p(\mathbf{E}_\mathrm{\gamma}) \cdot \overline{\mathbf{F}}_{\mathrm{\gamma}} \cdot \vec{\mu}_{\mathrm{tr}}\\
        \dot{\mathbf{K}}_{\mathrm{air}} &= \int {\overline{\mathbf{E}}_{\mathrm{\gamma}}^{\mathrm{tr}}} \cdot {\mathbf{\bar{E}}_{\mathrm{\gamma}}^{\mathrm{bin}}} \, de
    \end{aligned}
\end{equation}

Using this transformation, we obtain a 3D tensor \(\dot{\textbf{K}}_{\mathrm{air}}\) for the simulated radiation field that represents the relative air kerma per simulated photon. In order to retrieve an absolute measurand, like the air kerma rate, we need at least one reference measurement point to determine the linear conversion factor between the relative simulated air kerma rate and the real measurement so that simulated and measured kermas can be compared through all points. For our simulation validation, we decided to use the ratio between the integrals of simulated curve \(s(\vec{p}): \vec{p} \xrightarrow{} k\) and the measured curve \(m(\vec{p}): \vec{p} \xrightarrow{} k\) as the simulation conversion factor \(S_c\).

\begin{equation}
    \begin{aligned}
    S_c = \frac{\int m(\vec{p})\,d\vec{p}}{\int s(\vec{p})\,d\vec{p}}
    \end{aligned}
\end{equation}



\subsection{Experimental Validation}
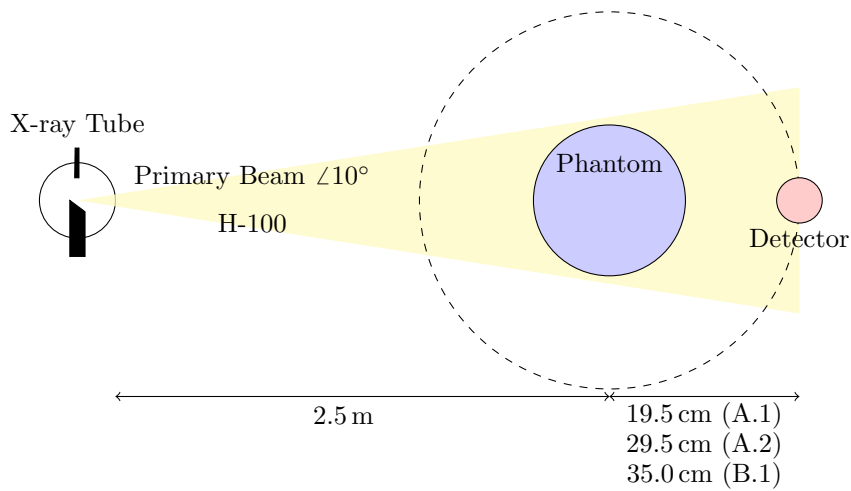
\begin{figure}[ht]
    \centering
    \begin{tikzpicture}
        \draw (-3.5, 0.5) circle (0.5);
        \filldraw (-3.6, -0.25) -- (-3.6, 0.5) -- (-3.4, 0.35) -- (-3.4, -0.25);
        \filldraw (-3.53, 1.2) -- (-3.536, 0.8) -- (-3.48, 0.8) -- (-3.48, 1.2);
        
        \node at (-3.5, 1.5) {X-ray Tube};
    
        \fill[yellow!30, opacity=0.75] (-3.5, 0.5) -- (6.0, 2.0) -- (6.0, -1.0) -- cycle;
        \node at (-1.2, 0.8) {Primary Beam $\angle 10^\circ$};
    
        \draw[fill=blue!20] (3.5, 0.5) circle (1);
        \node at (3.5, 1) {Phantom};
    
        \draw[dashed] (3.5, 0.5) circle (2.5);
    
        \draw[fill=red!20] (6, 0.5) circle (0.3);
        \node at (6, -0) {Detector};
    
        \draw[<->] (-3, -2.1) -- (3.5, -2.1);
        \node at (0, -2.35) {$2.5\,\mathrm{m}$};
    
        \draw[<->] (3.5, -2.1) -- (6, -2.1);
        \node at (4.75, -2.35) {$19.5\,\mathrm{cm}$ (A.1)};
        \node at (4.75, -2.75) {$29.5\,\mathrm{cm}$ (A.2)};
        \node at (4.75, -3.15) {$35.0\,\mathrm{cm}$ (B.1)};
    
        \node at (-1.2, 0.2) {H-100}; 
    \end{tikzpicture}
    \caption{Schematic top-down view of our measurements setup. The parameters for each of our measurement configuration, A and B, are 
    annotated at each value. Some parameters like the X-ray tube distance and the opening angle of the primary beam were fixed during 
    all measurements. The phantom was a cylindrical water barrel during configuration A that was exchanged with a male Alderson phantom 
    torso for configuration B.}
    \label{fig:ExperimentSetup}
\end{figure}

\begin{wrapfigure}{l}{5cm}
    \includegraphics[scale=0.1]{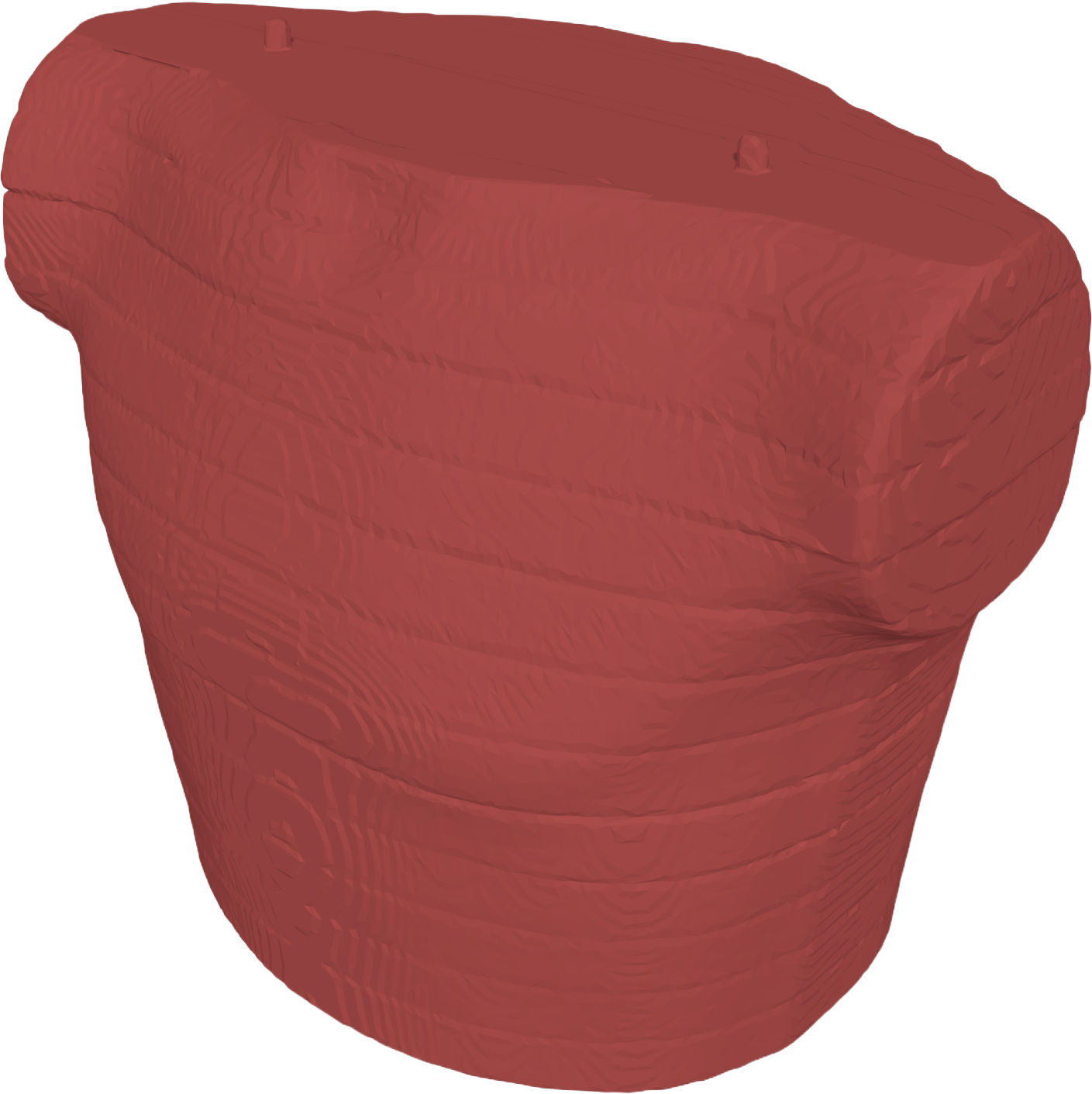}
    \caption{Render of the reconstructed \ac{CT}-scan of our male Alderson phantom torso as it was used in experiment B.}\label{fig:AldersonHeadless3DModel}
\end{wrapfigure}
For the experimental validation of our \ac{MCS} and to obtain the conversion factor \(S_c\) for a given configuration, we created a 
measurement setup in our laboratories using one of our X-ray facilities. The setup and all configuration parameters used are illustrated in \Cref{fig:ExperimentSetup}. In our laboratory, we created two configurations, A and B, that differ from each other by the type of scatter phantom. For configuration A, we used a simple water filled cylinder phantom with a diameter of \(20\,\mathrm{cm}\) and a height of \(20\,\mathrm{cm}\), that we modeled for our simulation in Blender~\cite{blender_online_community_blender_2024} by generating a cylindrical mesh. These phantoms are used to mimic the scatter behavior of a human head. For configuration B, we employed a more complex geometry as we used the torso of a male Alderson phantom which, as a whole, has a mass of \(73.5\,\mathrm{kg}\). For the simulation, we used a reconstructed 3D model of the irradiated volume of the Alderson phantom that we retrieved by performing a \ac{CT}-scan on it. The reconstruction was done semi-automatically using the software InVesalius~\cite{bebis_invesalius_2015} and the mesh reconstruct modifier of Blender. The reconstructed mesh can be seen in Figure~\ref{fig:AldersonHeadless3DModel}. We simplified the internal structure of the reconstructed mesh by filling the whole phantom with ICRU soft 4-component tissue according the ICRU-33 report~\cite{international_commission_on_radiation_units_and_measurements_icru_1980} except for the lung cavity, which we filled with ICRP lung tissue~\cite{international_commission_on_radiological_protection_icrp_2006}.

For our measurements, we placed the scatter phantom $2.5\,\mathrm{m}$ away from the X-ray tube. The phantom was then irradiated by an X-ray beam. In order to maximize the detector response in the scatter component, we used a H-100 radiation quality for all our experiments. The radiation field properties can be retrieved from the ISO 4037-1~\cite{iso_iso_2019}. Our X-ray tube creates a cone shaped primary beam. We used the PTW TK-30 type TM32005 -- a $30\,\mathrm{cm}^3$ spherical ionization chamber -- as our detector, which was the smallest spherical chamber available at our facility. In principle, during all the experiments, we measured in \(10\,\degree\) steps in a circle around the scatter phantom, except for one angle range per measurement that we could not use due to having to modify the experimental setup during the measurement series. Those ranges will be discussed in \replaced{the discussion section}{\Cref{sec:dis}}.

In configuration A, to measure the scattered, transmitted, and the combination of the direct X-ray beam and the scattered radiation field, we placed the water cylinder on a rotatable platform centered on the rotation axis and attached the detector successively at two distinct distances (\(19.5\,\mathrm{cm}\) and \(29.5\,\mathrm{cm}\)) from the water cylinder surface on the platform. 
We used two distances in order to assess the continuity and attenuation calculation of RadField3D. Using this platform, we were able to effectively rotate the detector in a circle around the water phantom to survey the present radiation field around the water phantom.

In configuration B, we were more interested in assessing the complex geometry handling of RadField3D and therefore we only measured the circle around the phantom at a single distance of \(35.0\,\mathrm{cm}\) from the surface. Additionally, we tested the simulated rotation of the radiation beam itself, as the human Alderson phantom is not rotation-invariant regarding the horizontal axis like the cylinder and, therefore, we needed to calculate a whole radiation field for each angle we measured. We placed the $30\,\mathrm{cm}^3$ spherical ionization chamber in a fixed position relative to the phantom surface at a \(45\,\degree\) angle to the sagittal plane at the height of the heart cavity and rotated the phantom around the horizontal axis. We simulated this by rotating the beam around the phantom, which is equivalent to rotating the phantom in a stationary beam. 

RadField3D was then executed using the corresponding parameters and meshes of each configuration and stored its data in our data format, Rad\textit{Filed}3D, which will be explained in the next section. For the validation dataset, we used a scoring volume centered around the phantom with the dimensions \(1\,\mathrm{m} \times 1\,\mathrm{m} \times 1\,\mathrm{m}\), meaning that the X-ray tube itself, as well as the first \(2\,\mathrm{m}\) of the primary beam cone, were excluded from the scoring volume. We simulated the \(p(E_\mathrm{\gamma})\) per voxel in the form of a normalized probability distribution with a resolution of \(32\) energy bins and an energy bin width of \(4.68\,\mathrm{keV}\), allowing for an energy range of up to \(150\,\mathrm{keV}\).
\added[R1C8]{The reasoning behind selecting this upper boundary was to encompass the expected tube energies found for general radiology and fluroscopy. According to the European Commission’s Radiation Protection Report No. 154~\cite{directorate-general_for_energy_and_transport_european_commission_now_known_as_european_2008}, these energies are between \(60\,\mathrm{keV}\) and \(120\,\mathrm{keV}\).} For the energy range of our validation, we found no advantage in using higher energy resolutions.



\subsection{Binary file format}
\begin{figure}[ht]
    \centering
    \includegraphics[width=\textwidth]{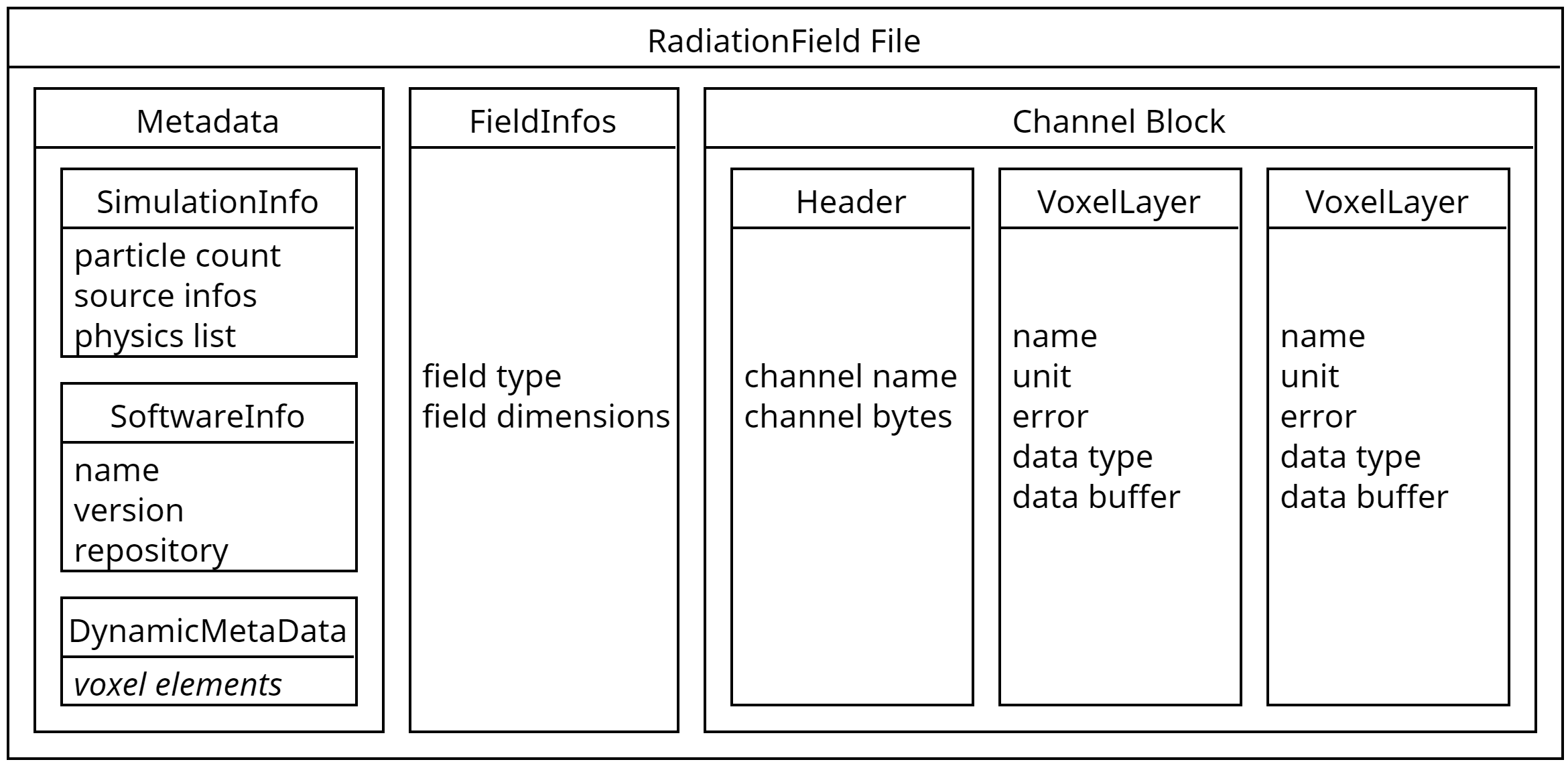}
    \caption{Schematic view of our Rad\textit{Filed}3D Format structure consisting of channel and layer blocks, preceded by a metadata 
    block. For better readability, the metadata block only shows the metadata elements partially. Please note, that channel and layer 
    blocks can be repeated.}
    \label{fig:FileFormatStructure}
\end{figure}

In order to evaluate and to archive the results of our simulations using RadField3D, we developed a minimalist binary file format. This file format should be fast in terms of read and write operations and easy to use. Further, it should preserve all relevant information about the configuration of the simulation that produced a single radiation field. Additionally, as the most popular language for data analysis and machine learning is Python, we created bindings for our C++ library so that we can access the binary file format from Python in an efficient way. For creating these bindings we used PyBind11~\cite{wenzel_pybind11_2016}.

In general, Rad\textit{Filed}3D stores one single radiation field with a given spatial resolution by defining voxel dimensions and an 
overall extent of the field. Each subsequently defined channel and layer will share the same dimensions and resolution. Each field is 
divided into channels which are just named containers for a set of layers. Each layer holds a specific simulated value per voxel. These 
values can be scalars of any native data type, but can also be vectors. Further, each layer is associated with a unit as well as a 
statistical error. Each layer is coherent and can be loaded and accessed independently of other layers. In order to preserve the configuration of the specific simulation, each binary file contains a header of metadata. There are two types of metadata: First, there are dynamic metadata which are just descriptions and single values structured like the data of a single voxel; Secondly, there are fixed metadata fields that adopt and extend the guidelines for documenting \ac{MCS}~\cite{sechopoulos_records_2018} and are mandatory.

The overall structure of our Rad\textit{Filed}3D format is illustrated in Figure~\ref{fig:FileFormatStructure}. For our specific requirements during the validation of RadField3D, we used two channel blocks to separate the primary beam and scatter component of the radiation field. Inside each channel, we stored two layer blocks: The \textit{photon energy distribution}, \(p(E_\mathrm{\gamma})\); and the \textit{amount of photon hits per primary photon}, \(\overline{F}_{\mathrm{\gamma}}\).

\section{Results}
\begin{figure}[h!]
    \centering
    \begin{minipage}[b]{0.45\textwidth}
        \centering
        \includegraphics[width=\textwidth]{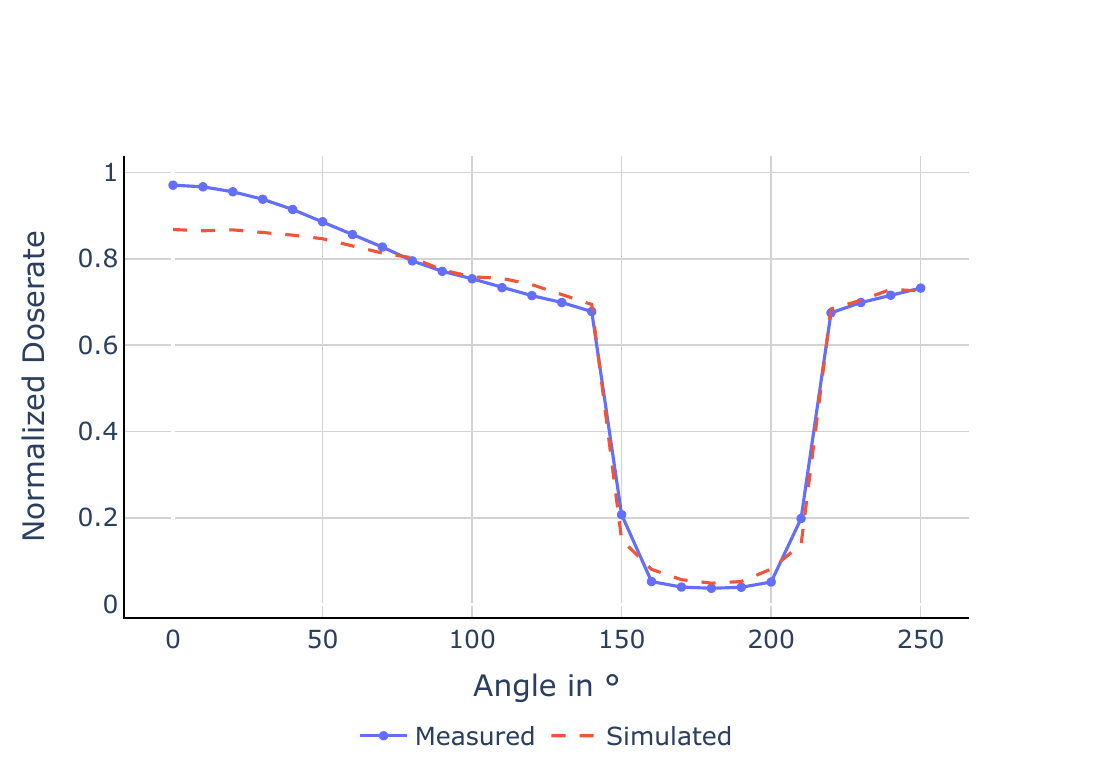}
        \caption{Experiment A.1: \(19.5\,\mathrm{cm}\) distance of the detector from the center of the water phantom.}\label{fig:simulation_comparisons:A.1}
    \end{minipage}
    \hfill
    \begin{minipage}[b]{0.45\textwidth}
        \centering
        \includegraphics[width=\textwidth]{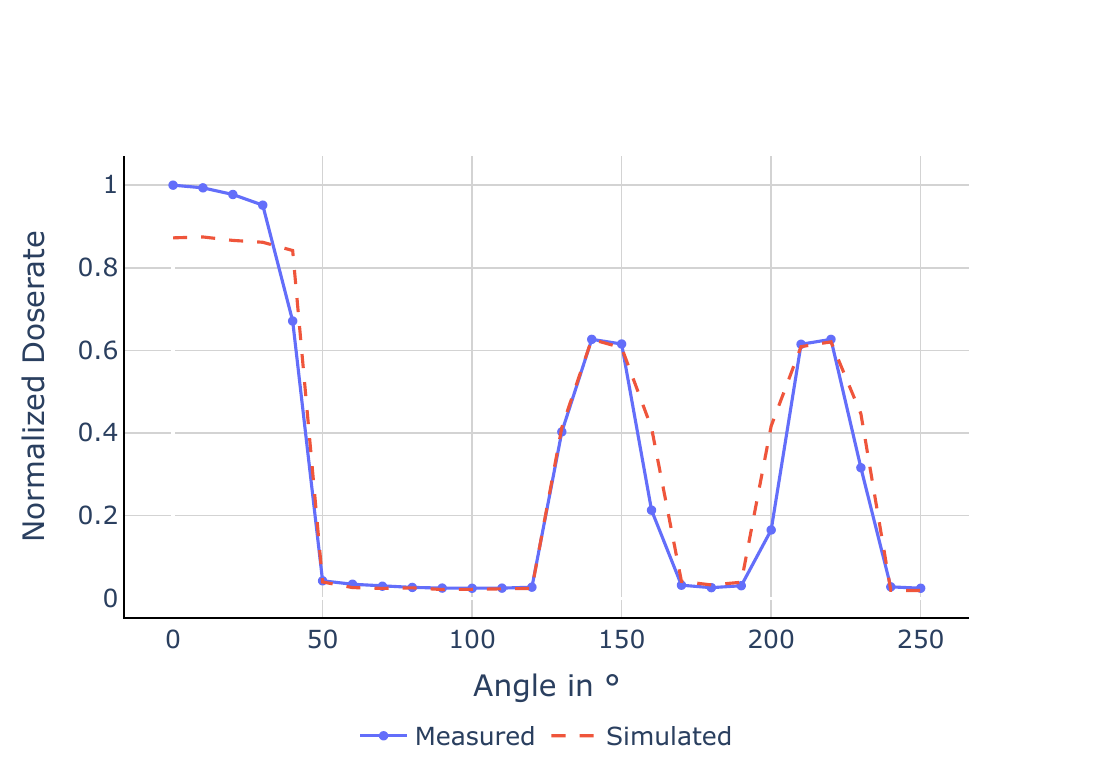}
        \caption{Experiment A.2: \(29.5\,\mathrm{cm}\) distance of the detector from the center of the water phantom.}\label{fig:simulation_comparisons:A.2}
    \end{minipage}
    \vfill
    \begin{minipage}[b]{0.45\textwidth}
        \centering
        \includegraphics[width=\textwidth]{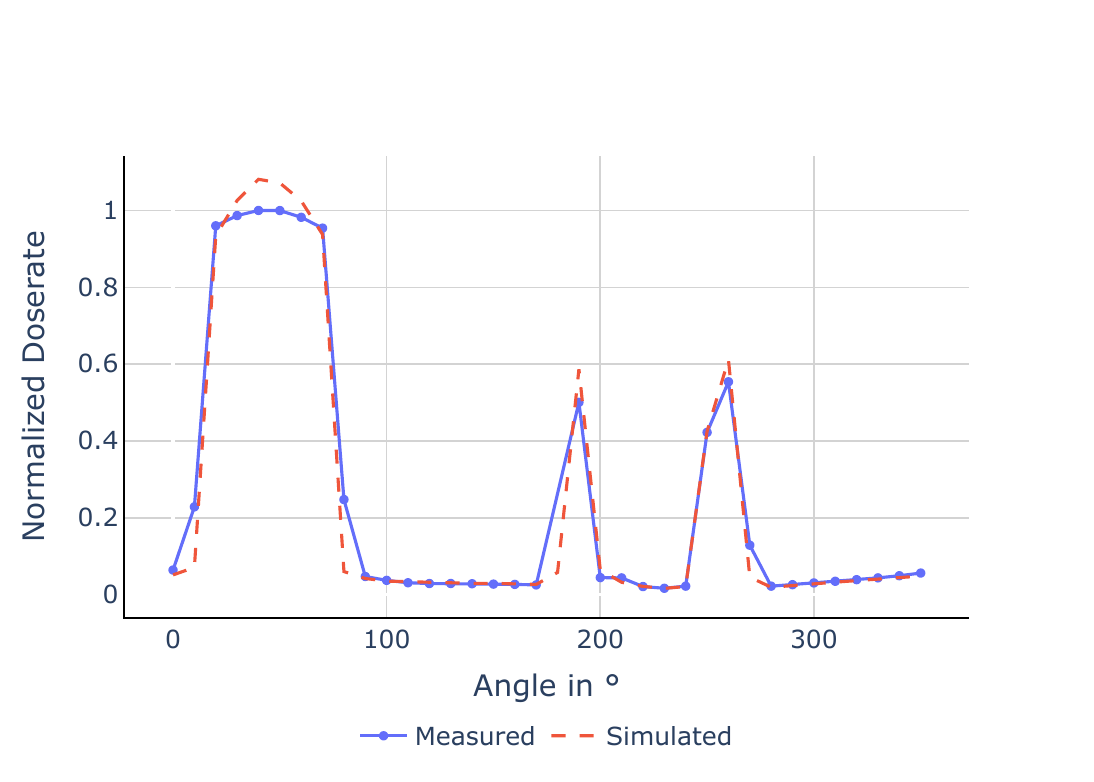}
        \caption{Experiment B.1: \(35.0\,\mathrm{cm}\) distance of the detector from the center of the male Alderson phantom at heart cavity height.}\label{fig:simulation_comparisons:B}
    \end{minipage}
    \hfill
    \begin{minipage}[b]{0.45\textwidth}
        \centering
        \includegraphics[width=\textwidth]{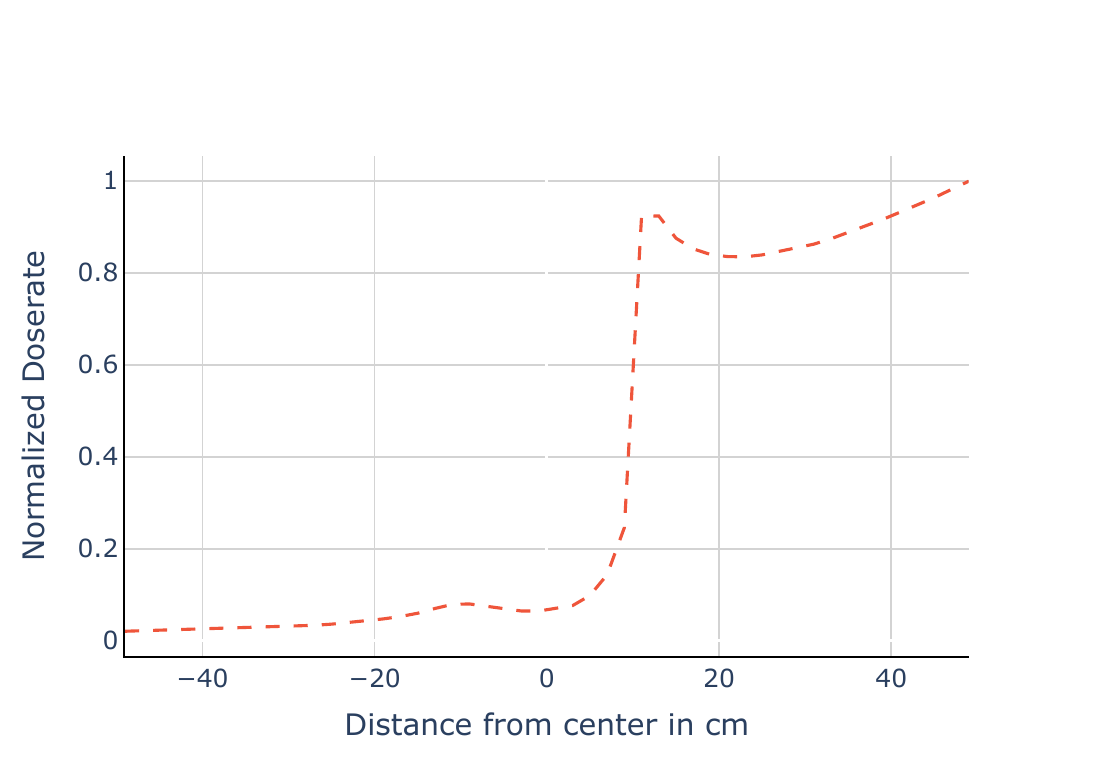}
        \caption{Simulation of configuration A: \replaced[R2C4]{Decease}{Decrease} of the relative air kerma rate along the center line in the xy-plane.}\label{fig:simulation_decease:A}
    \end{minipage}
    \vfill
\end{figure}

We assessed the quality of our simulation and therefore the quality of the possibly created datasets from our application RadField3D by comparing the relative simulation results against relative air kerma rates \(\dot{K}_{\mathrm{air}}\) measured with a spherical ionization chamber as described prior. That allowed us to later calibrate our simulation results to one or more known measurement points to deduce absolute simulation values for different positions from it.

At first, we wanted to ensure that our simulation can handle simple geometry such as a water cylinder (\replaced[]{A}{a} human head phantom). Therefore, we placed the phantom on a rotatable plate \(2.5\,\mathrm{m}\) in front of an X-ray tube and irradiated it. For practical reasons we were unable to rotate the detector higher than \(250\,\degree\) in the first configuration. This is referred to as configuration A in the following and was used for the experiments A.1 and A.2, whose results are shown in \Cref{fig:simulation_comparisons:A.1,fig:simulation_comparisons:A.2}. For experiment A.1, where the detector was \(19.5\,\mathrm{cm}\) away from the rotation axis, we found a good correspondence of the simulated curve with the measured points. That allowed us to conclude that the simulation application correctly handles occlusion of the X-ray beam by a scatter object.
\replaced[R2C2]{In general, all experiments demonstrated a good geometrical agreement as well as a sufficient agreement regarding the relative air kerma rates \(\dot{K}_{\mathrm{air}}\).}{In general, all experiments demonstrated a good geometrical agreement. Additionally, we observed a sufficient agreement regarding the relative air kerma rates \(\dot{K}_{\mathrm{air}}\) according to the feasibility study performed by U. O'Connor~\etal~\cite{oconnor_feasibility_2022}, which concluded that a deviation of up to \(40\,\%\) regarding \(\mathbf{H}_\mathrm{p}(10)\) is acceptable for a computational dosimetry system. As we score photon radiation only, the relationship between \(\dot{K}_{\mathrm{air}}\) and \(\mathbf{H}_\mathrm{p}(10)\) is defined by an approximately constant linear factor for our energy range and, thereby, we can use that as a guidance.}

During configuration A.2, we aimed to measure more of the overlap of the scattered radiation and the primary X-ray beam and compare this to our simulation. Therefore, we increased the distance of the detector from the rotation axis to \(29.5\,\mathrm{cm}\). The results of that are shown in \Cref{fig:simulation_comparisons:A.2}. At this distance, we also found a good correspondence of both curves. As it can be observed in \Cref{fig:simulation_decease:A}, the simulated radiation field showed the expected one over square drop in undisturbed air volumes, therefore, we assume, that this validation should hold for the whole radiation field.


In order to test our simulation also for more realistic configurations and especially complex geometry, in configuration B we used the male Alderson phantom torso and the reconstructed 3D mesh of it for the measurement and simulation. Here, we also found a good geometrical accordance. However, the simulated \(\dot{K}_{\mathrm{air}}\) at the local maxima tends to overestimate the measurements. This can be seen in \Cref{fig:simulation_comparisons:B}. As the detector was fixed at \(45\degree\) relative to the Alderson phantom surface, the curve of experiment B.1 is shifted by this angle with respect to the experiments of the configuration A.

\begin{table}[htb]
    \centering
    \setlength\tabcolsep{15pt}
    \begin{tabular}{ l | c | c | c }
     Experiment & \(\tilde{e}_{rel}\) & \(\overline{e}_{rel}\) & \(\sigma(e_{rel})\) \\
     \midrule
     A.1 & \(4.0\,\%\) & \(13.5\,\%\) & \(17.2\,\%\) \\
     A.1 (No Edges) & \(3.6\,\%\) & \(10.9\,\%\) & \(13.1\,\%\) \\
     A.2 & \(8.9\,\%\) & \(18.5\,\%\) & \(24.9\,\%\) \\
     A.2 (No Edges) & \(8.0\,\%\) & \(11.9\,\%\) & \(15.6\,\%\) \\
     B.1 & \(6.6\,\%\) & \(16.0\,\%\) & \(21.6\,\%\) \\
     B.1 (No Edges) & \(6.4\,\%\) & \(7.8\,\%\) & \(5.5\,\%\) \\
    \end{tabular}
    \caption{List of the relative point error \(e_{rel}\) distribution descriptors for each experiment: Median~(\(\tilde{e}_{rel}\)), Mean~(\(\overline{e}_{rel}\)) and Standard Deviation~(\(\sigma(e_{rel})\)). The annotation "No Edges" refers to the exclusion of the angles at which a sudden change inside the radiation field occurs.}
    \label{tab:experimentErrors}
\end{table}

In order to describe the comparisons in a quantitative manner, \Cref{tab:experimentErrors} contains the median and mean relative errors for each experiment under different evaluations. Therefore, we computed the point-wise relative error \(e_{rel}\) by dividing the absolute deviation of the measured and corresponding simulated values by the measured value. Additionally, we provide the standard deviation of the series of point-wise relative errors.
\section{Discussion}\label{sec:dis}
In our experiments, we were able to show that our \ac{MCS} application creates spatially resolved radiation fields whose relative topology could be proved to be reasonable. We demonstrated that the trajectory of the air kerma rate \(\dot{K}_{\mathrm{air}}\) is uninterrupted across the field, as shown in~\Cref{fig:simulation_decease:A}. This is significant in the context of our voxelization, as this is not conducted analytically by calculating the intersection points between cubic voxels and photon paths. Rather, it is accomplished by testing at discrete points along the photon path. As a result, voxels that are only slightly intersected are not detected. However, due to the probabilistic nature of a \ac{MCS}, this simplification has a negligible impact on the outcome. Moreover, our findings demonstrate that the data of a radiation field can be employed to estimate the air kerma rate \(\dot{K}_{\mathrm{air}}\) at any position within the voxel grid, provided that at least one point is known to be located within the range of the simulated radiation field. In regard to the comparison of the \(\dot{K}_{\mathrm{air}}\) values across all experiments, it was demonstrated that the point-wise relative error of the measurement and the simulation, when taken the median, is well below \(10\,\%\) according to \Cref{tab:experimentErrors}. Concurrently, the mean of those errors is markedly elevated, reaching as high as \(18.5\,\%\) in experiment A.2. This is attributable to the elevated errors observed at hard edges inside of the radiation field, induced by the geometry and the primary beam. For experiment A, this encompassed the angles \(130\degree\), \(160\degree\), \(200\degree\) and \(230\degree\) for the geometry induced edges and \(40\degree\) for the edge caused by the primary beam. Upon exclusion of these angles for measurement A.2, the mean error declines to \(11.9\,\%\), with a standard deviation of \(15.6\,\%\) and a median of \(8\,\%\). This trend applies to the other two measurements as well. For experiment B, we excluded the angles \(10\degree\) and \(80\degree\) as the edges of the primary beam and the angles \(180\degree\), \(200\degree\), \(250\degree\), and \(270\degree\) as the edges of the Alderson phantom. The high errors at the edges of a hard change inside the radiation field can be explained with the spatial extent of the ionization chamber in combination with the voxelization of the simulated radiation field.

\added[R1C1, R1C9, R2C3]{In order to simplify the validation section and to present our data as honest as possible, we presented our simulation results raw as our files are containing them. The main reasons for high differences between the simulation and the experiments are the voxelization error and positioning uncertainty during measurements. That influences locations with steep gradients between the adjacent voxels the most. A way to correct for this effect is to blend between neighboring voxels. We were able to roughly cut the standard deviation in half for experiments A.2 and B.1 by using the following approach: we take the assumed center of our detector and calculate the distance between all neighboring voxel centers with a cut-off distance of the detector radius; then, we blend the fluence and spectra linearly based on the distance from the detector center. That way, we achieved a standard deviation of \(13.4\,\%\) for experiment A.2, and \(11.2\,\%\) for B.1; over all relative errors with a mean of \(12.9\,\%\) for A.2, and \(11.2\,\%\) for B.1, without excluding the angles with hard edges in the radiation field. Further, the results can be improved by splitting the radiation field in two components: the primary beam, and the scattered component. Therefore, \textit{RadField3D} already outputs the results split into these two component which were combined for all previous results. When just comparing the regions with similar absolute fluences like primary beam, scattered field, and transmitted beam separately from each other, we observe relative error means ranging from \(1.8\,\%\) to \(5.5\,\%\) across all three experiments.}

All in all, the experiments demonstrate, that the major absolute \(\dot{K}_{\mathrm{air}}\) discrepancies can be observed within the primary beam component. Nevertheless, for the case of \ac{IR}, this component is already well known without the use of \ac{MCS}, as the beam is measured continuously. However, for the scatter component, we observe a slight tendency to overestimate the measured \(\dot{K}_{\mathrm{air}}\). \replaced[R2C5]{This behavior is not a huge deal in the context of radiation protection as we aim to retrieve an upper limit for the received doses of the monitored personnel.}{This behavior, however, is not a significant concern within the context of radiation protection, as our primary objective is to establish an upper limit for the doses received by the monitored personnel.}

Aside from the quality of the \ac{MCS} application itself, we have developed a file format that allows for fast and structured loading of datasets. We believe that the existence of a domain specific data format for radiation fields will improve the interoperability of researchers using different \ac{MCS} applications especially, but not limited to, in the context of machine learning or deep learning on dosimetric data.

\section{Conclusion}
We have presented \textbf{RadField3D}, a validated radiation transport simulation application based on the Geant4 framework.
With the Geant4 simulation framework and our voxelization method, we demonstrated RadField3D to be capable of generating spatially resolved three-dimensional radiation fields with an error that is reasonable. For reproducibility and reusability purposes, we additionally presented the corresponding data format \textbf{Rad\textit{Filed}3D}, which is characterized by its rapid processing and machine-interpretable capabilities. Its Python \ac{API} allows the smooth integration with existing machine and deep learning frameworks. This will be useful for future research in the field of dosimetry, e.g., the research of novel simulation methods aside from \ac{MCS}.

\deleted{All our data and source code will be made publicly available upon acceptance.} We believe that our work is highly useful to the community and opens many opportunities for future applications and further research.


%
%
%
%




\section*{Acknowledgments}
%
The authors gratefully acknowledge partial funding by the DFG under Germany’s Excellence Strategy within the Cluster of Excellence PhoenixD (EXC 2122, Project ID 390833453) and a DFG Research Grant (HU 2660/3-1, "Development of real-time capable methods for the simulation of photon radiation - using the example of quantitative dosimetry in interventional radiology", Project number 547148940).


\bibliographystyle{unsrt}

\bibliography{bibliography-formatted}

\end{document}